\documentclass[runningheads]{llncs}

\RequirePackage{silence}
\usepackage[T1]{fontenc}

\usepackage{silence}
\usepackage{multirow}
\usepackage{graphicx}
\makeatletter
\let\vec\@undefined
\makeatother
\usepackage{amsmath,amssymb}
\usepackage{booktabs}
\usepackage{cite}
\usepackage{float}
\usepackage{microtype} 
\usepackage{xcolor} 
\definecolor{mygreen}{rgb}{0.0, 0.5, 0.0} 
\usepackage{tcolorbox} 

\begin{document}

\title{CrossSpine: Multi-scale Cross-sequence Attention with Anatomical Priors for Automated Pfirrmann Grading}

\titlerunning{CrossSpine: Multi-scale Cross-sequence Attention}

\author{Hai Son Nguyen\inst{1} \and
Duong Ngoc Vu\inst{1} \and
Trong-Nghia Nguyen\inst{1} \and
Bien Tran Van\inst{3,4} \and Van-Dem Pham\inst{5} \and
Trang Mai Xuan\inst{2}\thanks{Corresponding author.} \and Huan Vu\inst{1} \and Thien Van Luong\inst{1}}

\authorrunning{H. S. Nguyen \textit{et al.}}

\institute{Business AI Lab, College of Technology, National Economics University, Vietnam\\
\and
A2I Lab, Phenikaa School of Computing, Phenikaa University, Hanoi, Vietnam\\
\and
Medical Imaging \& Radiological Technology Department, Faculty of Medical Technology, Phenikaa School of Medicine \& Pharmacy, Phenikaa University, Vietnam\\
\and
Radiology \& Functional Exploration Center, Phenikaa University Hospital, Vietnam
\and
Deparment of Pediatrics, Hospital of University Medicine and Pharmacy, Vietnam National University, Hanoi, Vietnam
\\
\email{hains24206@gmail.com, duongvn.bai@st.neu.edu.vn, nghiant@neu.edu.vn, bien.tranvan@phenikaa-uni.edu.vn, dempv.ump@vnu.edu.vn, trang.maixuan@phenikaa-uni.edu.vn, \{huanv, thienlv\}@neu.edu.vn}
}
\maketitle            
\begin{abstract}
Automated grading of Lumbar Disc Degeneration is essential for the objective quantification of structural changes associated with low back pain. Observing that baseline models underperformed on our data, we propose a framework designed to overcome these limitations. First, we present the Cross-sequence Attention Spine (CrossSpine) framework, a novel architecture that employs a cross-sequence attention mechanism to adaptively fuse features from different MRI sequences at multiple spatial scales. Second, we contribute a meticulously curated dataset aimed at automated Pfirrmann grading. Finally, we introduce an IVD-aware classification technique that integrates anatomical disc-level information, enabling the model to learn level-specific degeneration priors. Our experiments demonstrate the superiority of this approach: CrossSpine achieved a relative improvement exceeding 125\% in the Macro F1 score, while boosting the Mean AUPRC by 99\% and the Mean AUROC by 36\% compared to the baseline.

\end{abstract}

\section{Introduction}

Lumbar Disc Degeneration (LDD) has been identified as one
of the several causes of Low Back Pain~\cite{hartvigsen2018low}. In clinical practice, Magnetic Resonance Imaging (MRI) is the standard modality for diagnosis, typically using the Pfirrmann grading system to quantify degeneration based on T2-weighted images~\cite{pfirrmann2001magnetic}. 
While deep learning models have achieved promising results in healthcare diagnostics~\cite{nguyen2025temporal, asif2025advancements}, the performance of baseline models is insufficient owing to two key issues. Primarily, their design for single-stream processing prevents the effective integration of complementary features from different MRI sequences. Additionally, the uniform treatment of all lumbar discs overlooks the specific anatomical context and degeneration patterns distinct to each IVD level.

We observed that baseline models underperformed on our dataset. To address this limitation, we propose a framework designed to overcome this challenge and fully leverage the potential of our data. Our main contributions are summarized as follows: (1) We present a novel CrossSpine architecture that employs a cross-sequence attention mechanism to fuse features from different MRI sequences at multiple scales, thereby improving the model's ability to capture subtle degenerative changes that may only be visible in specific sequences. Through the self-attention mechanism proposed by Vaswani et al.~\cite{vaswani2017attention}, each sequence can reference and learn complementary information from all other sequences, enabling mutual feature enhancement. Furthermore, after this inter-sequence learning, we introduce a sequence weighting mechanism that automatically identifies and prioritizes the most diagnostically informative sequences. (2) We present a dataset designed to support automated Pfirrmann grade classification. (3) We introduce an IVD-aware classification technique that explicitly integrates anatomical disc-level information into the model, enabling the model to learn level-specific degeneration priors. Crucially, we demonstrate that this technique is model-agnostic: our experiments show that injecting disc level consistently improves the performance of both our proposed architecture and baseline models.

\section{Related Work}
In recent years, the application of deep learning techniques for the processing and diagnosis of MRI has become increasingly prevalent~\cite{lundervold2019overview}. In this study, we utilized deep learning architectures as baseline models for Pfirrmann grading. These included established convolutional networks such as ResNet introduced by He et al.~\cite{he2016deep}, DenseNet by Huang et al.~\cite{huang2017densely}, and EfficientNet by Tan et al.~\cite{tan2019efficientnet}, as well as modern transformer-based models like ViT-Base proposed by Dosovitskiy et al.~\cite{dosovitskiy2020image}. Despite their proven efficacy in general visual recognition, these models exhibited poor feature representation capabilities when applied to our data. We attribute this underperformance to the lack of specialized feature fusion mechanisms within these backbones, which prevents them from effectively learning information across multiple MRI sequences. This observation highlights the necessity for a dedicated architecture capable of leveraging the rich, complementary information provided by our dataset. To address this, we introduce an attention mechanism that enables the model to process multiple sequences jointly. By leveraging the self-attention strategy formulated by Vaswani et al.~\cite{vaswani2017attention}, our architecture allows different MRI contrasts to mutually reference one another, rather than treating them in isolation. This facilitates dynamic feature selection, where the network automatically learns to assign higher importance to the most relevant sequences.

To the best of our knowledge, most existing state-of-the-art methods~\cite{natalia2024lumbar, kowlagi2023stronger} do not explicitly use the anatomical position of the intervertebral discs in their models. They treat all discs identically, ignoring the specific spinal level. This oversight is critical because degeneration patterns are biomechanically distinct across different levels. Our work aims to bridge these gaps by proposing a specialized architecture that outperforms backbones on our data, supported by a uniquely diverse dataset and novel anatomical integration.

\section{Materials and Methods}
\begin{figure}[H] 
    \centering
    \includegraphics[width=\linewidth, trim=0 730 0 0, clip]{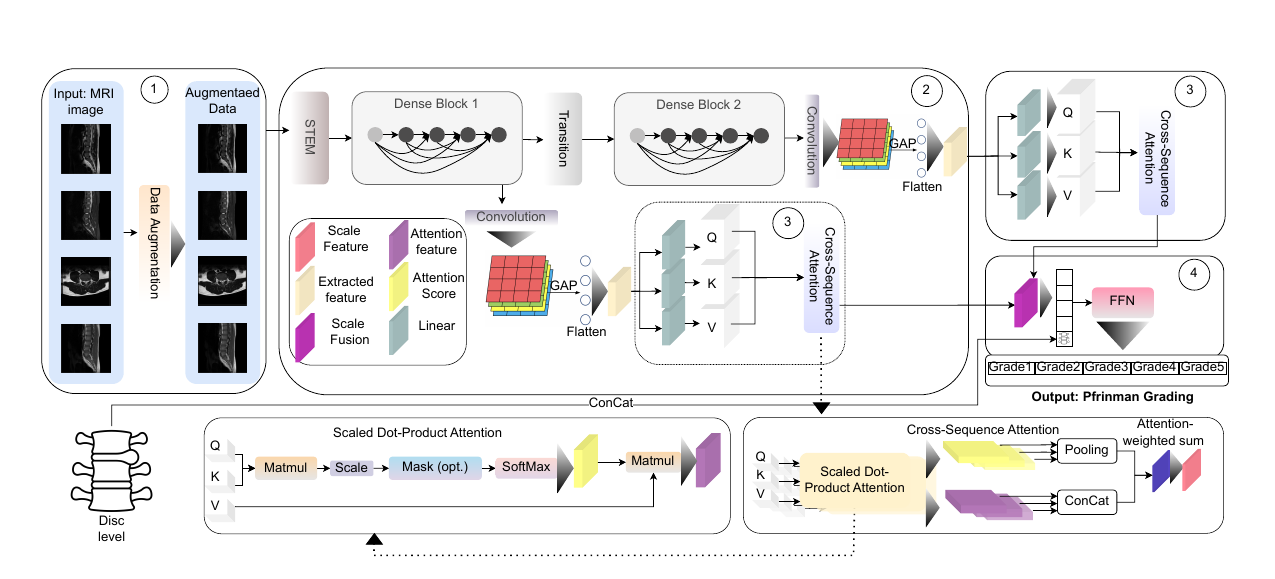} 
    \caption{Overview of CrossSpine for MRI Spine Analysis. Our system consists of 4 steps: (1) Data Preprocessing, (2) Feature Extraction, (3) Attention Fusion, (4) IVD-aware classification.}
    \label{fig:pipeline_arch}
\end{figure}
\begin{table}[htbp]
  \centering
  \caption{Pfirrmann Grade Distribution of PhenMec.}
  \label{tab:phenmec_distribution}
  \setlength{\tabcolsep}{10pt} 
  \begin{tabular}{lrr}
    \toprule
    \textbf{Pfirrmann Grade} & \textbf{Count} & \textbf{Percentage} \\
    \midrule
    Grade 1 & 35 & 3.0\% \\
    Grade 2 & 663 & 56.0\% \\
    Grade 3 & 400 & 33.7\% \\
    Grade 4 & 83 & 7.0\% \\
    Grade 5 & 4 & 0.3\% \\
    \midrule
    \textbf{Total} & \textbf{1,185} & \textbf{100.0\%} \\
    \bottomrule
  \end{tabular}
\end{table}
\subsection{Dataset}
We introduce a multi-view MRI benchmark for comprehensive spine analysis. As shown in Table~\ref{tab:phenmec_distribution}, Grade~V samples are extremely rare in our data.
The dataset was collected at PhenMec Hospital from July to September 2025, yielding a total of 16,813~raw DICOM images across sagittal and axial planes. After applying inclusion criteria, the final curated dataset comprises 1,185~IVD records derived from 237~unique patients. Structurally, the data achieves anatomical balance, containing exactly 237~instances for each intervertebral disc level from L1--L2 to L5--S1. The Pfirrmann grade distribution mirrors the inherent imbalance of clinical environments, characterized by a dominant Grade~II class and a marked scarcity of cases in Grades~I and~V.

\subsection{Data Preprocessing}
\subsubsection{DICOM Processing.}
Our input data consists of clinical MRI examinations stored in DICOM format. Each examination includes different imaging sequences, and each sequence contains multiple images of the lumbar~spine. Among the images within each sequence, we observe that the middle image provides the sharpest and clearest visualization of the entire lumbar spine. Therefore, for each sequence, we select the middle image as the input for our model and resize all images to $224 \times 224$ pixels. To fully utilize the available data and capture complementary diagnostic information, we use all four sequences as inputs to our model rather than relying on a single sequence.

\subsubsection{Data Augmentation.}
We use data augmentation to generate more training examples. These augmentations create variations of the training images that facilitate the learning of robust features, thereby improving model generalization and preventing overfitting.
The augmentation pipeline includes random horizontal flipping, random rotation up to 15 degrees, translation up to $10\%$ of the image size, scaling between $90\%$ and $110\%$, shearing up to $5^\circ$, and brightness and contrast jittering up to $20\%$.

\subsection{Feature Extraction}
The second stage extracts features from the preprocessed images. A key innovation of our approach is that we extract features at two different spatial scales, rather than using only the final network output.

\subsubsection{Stem Layer.}
We employ DenseNet-121, developed by Huang et al.~\cite{huang2017densely}, as our backbone for feature extraction. To facilitate representation learning, the network is initialized with weights pretrained on the ImageNet~\cite{russakovsky2015imagenet} dataset. Input images are processed through the first convolution layer, which is modified to accept single-channel input; when pretrained weights are used, we initialize it by averaging the original RGB weights across channels. Each input image passes through a stem layer that performs initial feature processing. The stem consists of three sequential operations. First, a $7 \times 7$ convolution with stride 2 reduces the spatial resolution from $224 \times 224$ to $112 \times 112$ while expanding the single grayscale channel to 64 feature channels. Second, batch normalization followed by ReLU activation stabilizes the feature distribution and introduces non-linearity. Third, a $3 \times 3$ max pooling operation with stride 2 further reduces the spatial dimensions to $56 \times 56$.

\subsubsection{Multi-Scale Feature Extraction.}
The first scale extracts features after the first dense block. The feature maps have 256 channels with spatial size $56 \times 56$. We apply a convolution layer to reduce the channel dimension from 256 to 128, followed by batch normalization for training stability. Then, global average pooling (GAP) computes the mean value across all spatial positions for each channel, converting the $56 \times 56 \times 128$ feature maps into a 128-dimensional vector. This operation removes spatial information and produces a compact representation suitable for the attention mechanism~\cite{vaswani2017attention}. The second scale extracts features after the second dense block. The feature maps have 512 channels with spatial size $28 \times 28$. We apply the same processing pipeline: a convolution layer reduces channels from 512 to 256, batch normalization stabilizes the features, and global average pooling converts the $28 \times 28 \times 256$ feature maps into a 256-dimensional vector. The two scales thus produce feature vectors of dimensions 128 and 256 respectively, which are then passed to the cross-sequence attention stage.

\subsection{Attention Fusion}
We obtain feature vectors from four different sequences at each of two scales. Instead of concatenating these features, our model uses the cross-sequence attention mechanism inspired by Vaswani et al.~\cite{vaswani2017attention} to adaptively learn which sequences are most informative.

\subsubsection{Scaled Dot-Product Attention.}
Before computing the attention scores, we first project the extracted features into the query, key, and value spaces. Let $\mathbf{X} \in \mathbb{R}^{N \times d_{\text{model}}}$ denote the matrix of stacked feature vectors from the $N$ sequences (where $N=4$) obtained from the backbone at a specific scale. We apply three separate learnable linear layers to $\mathbf{X}$ to generate the Query ($\mathbf{Q}$), Key ($\mathbf{K}$), and Value ($\mathbf{V}$) matrices:
\begin{equation}
\mathbf{Q} = \mathbf{X} \mathbf{W}^Q, \quad \mathbf{K} = \mathbf{X} \mathbf{W}^K, \quad \mathbf{V} = \mathbf{X} \mathbf{W}^V, 
\end{equation}
where $\mathbf{W}^Q, \mathbf{W}^K, \text{and } \mathbf{W}^V$ are the weight matrices of the linear projections.

The attention output is then computed using these generated matrices as:
\begin{equation}
\text{Attention}(\mathbf{Q}, \mathbf{K}, \mathbf{V}) = \text{softmax}\left(\frac{\mathbf{Q}\mathbf{K}^T}{\sqrt{d_k}}\right)\mathbf{V}, 
\end{equation}
where $d_k$ is the dimension of the key vectors. The $\frac{1}{\sqrt{d_k}}$ serves as a scaling factor.
\subsubsection{Multi-Head Attention.}
In transformer~\cite{vaswani2017attention} architectures, multi-head attention often involves additional projections within each head. In our implementation, since we have already projected the features into $\mathbf{Q}, \mathbf{K}, \mathbf{V}$ as described above, we omit further projections to reduce computational complexity. We directly partition the feature dimension $d_{\text{model}}$ into $h$ parallel heads.
The matrices $\mathbf{Q}, \mathbf{K}, \text{and } \mathbf{V}$ are split into $h$ segments, where each segment corresponds to a subspace of dimension $d_k = d_{\text{model}}/h$. Let $\mathbf{Q}_i, \mathbf{K}_i, \mathbf{V}_i$ denote the $i$-th segment of the query, key, and value matrices, respectively. For each head $i$, the attention is computed as:
\begin{equation}
\text{head}_i = \text{Attention}(\mathbf{Q}_i, \mathbf{K}_i, \mathbf{V}_i). 
\end{equation}
The outputs of all heads are then concatenated to restore the original feature dimension:
\begin{equation}
\text{MultiHead}(\mathbf{Q}, \mathbf{K}, \mathbf{V}) = \text{Concat}(\text{head}_1, \ldots, \text{head}_h). 
\end{equation}
In our implementation, we use $h=4$ heads and set $d_k = d_v = d_{\text{model}}/h$.

\subsubsection{Cross Attention Sequence Fusion.}
We obtain feature vectors for each sequence, containing information integrated from all other sequences. To produce a fused feature vector for classification, we compute a weighted sum of the attended sequence features, where the weights are derived from the attention scores averaged across all heads and all query positions. For the first scale, this process yields a single feature vector of length 128. For the second scale, it produces a vector of length 256.

\subsection{Scale Fusion and IVD-Aware Classification}
A key contribution of this stage is the incorporation of anatomical disc level information, which we call IVD-aware classification.

\subsubsection{Multi-Scale Fusion.}
We have two fused feature vectors: a 128-dimensional vector from the first scale and a 256-dimensional vector from the second scale. To combine these, we first project both vectors to a common 512-dimensional space using separate linear transformation layers, then combine the projected features using learnable scale weights. These weights determine how much each scale contributes to the final representation. The weights are initialized to equal values and updated during training. After fusion, we apply layer normalization to stabilize the feature distribution.

\subsubsection{IVD-Aware Classification.}
Different lumbar disc levels exhibit distinct degeneration patterns and baseline rates. Clinical studies have shown that lower lumbar discs experience greater mechanical stress and show higher rates of degeneration compared to upper levels~\cite{adams2006intervertebral}. Additionally, the visual appearance of degeneration may vary across levels due to anatomical differences in disc size, shape, and surrounding structures. To leverage these level-specific priors, we design a classification process. First, a multi-layer perceptron (MLP) projects the 512-dimensional fused features into a compact 10-dimensional representation. Second, we explicitly incorporate the IVD level information by concatenating it with the 10-dimensional vector to form an 11-dimensional vector. This combined vector is passed through a final linear layer to produce the output grading. This late-stage integration allows the model to adjust its predictions based on the specific disc level without interfering with the preceding visual feature extraction.
For the CrossSpine without IVD variant, we use the same multi-scale feature extraction, cross-sequence attention, and scale-fusion pipeline to obtain the 512-dimensional fused representation, without the disc-level input. The final prediction is produced by an MLP head that maps the fused features directly to the 5 Pfirrmann grade.

\subsubsection{Model-Agnostic Contribution.}
Importantly, our IVD-aware technique is model-agnostic and can be applied to any classification architecture. We demonstrate experimentally that adding IVD level information improves both our model and simpler baseline architectures. This general applicability makes the technique a standalone contribution for spine imaging research.

\subsubsection{Loss Function.}We minimized a weighted cross-entropy objective function, as described by Goodfellow et al.~\cite{goodfellow2016deep}, defined as:
\begin{equation}
    \mathcal{L} = -\sum_{c=1}^{C} w_c \cdot y_c \log\left(\frac{\exp(\hat{y}_c)}{\sum_{j=1}^{C}\exp(\hat{y}_j)}\right),
\end{equation}
where $y_c$ denotes the ground truth label, $\hat{y}_c$ represents the predicted logit for class $c$, and $w_c$ is the class weight derived from the inverse frequency distribution adapted from Eigen et al.~\cite{eigen2015predicting} with a smoothing factor $\gamma$:
\begin{equation}
    w_c = \frac{N}{C \cdot n_c + \gamma \cdot N},
\end{equation}
$N$ represents the total number of training samples and $n_c$ is the sample count for class $c$.

\subsubsection{Missing Sequence Handling.} When certain MRI sequences are unavailable for a sample, zero-padding is applied to maintain consistent tensor dimensions. An availability mask is propagated through the attention mechanism~\cite{vaswani2017attention} to prevent information leakage from missing sequences.
\section{Experiments}
\subsection{Experimental Setup}
\subsubsection{Implementation.}
The dataset was chronologically partitioned into training ($70\%$), validation ($15\%$), and testing ($15\%$) subsets. All experiments were conducted on an NVIDIA RTX 4090 GPU. To address class imbalance and provide a comprehensive evaluation of classification performance, we prioritized Macro F1-score, AUROC, and AUPRC, as these metrics provide a more robust evaluation of minority classes.
\subsubsection{Baseline Architecture.}
To benchmark our proposed method, we implement a baseline model using DenseNet-121~\cite{huang2017densely} pretrained on ImageNet. Each of the four MRI sequences is processed independently, yielding a 1024-dimensional embedding via global average pooling. These embeddings are concatenated into a 4096-dimensional vector and classified using MLP. This classifier sequentially reduces the feature dimension to 256 and 128, before finally mapping to 5 output logits corresponding to the Pfirrmann grades. For the IVD-aware variant, the MLP is modified to output a 10-dimensional vector, which is subsequently concatenated with a disc-level indicator before the final linear projection. This architecture represents a standard fusion strategy lacking multi-scale processing or cross-sequence interaction.
\subsection{Experimental Results}
\begin{figure}[H]
    \centering
    \includegraphics[width=1\linewidth]{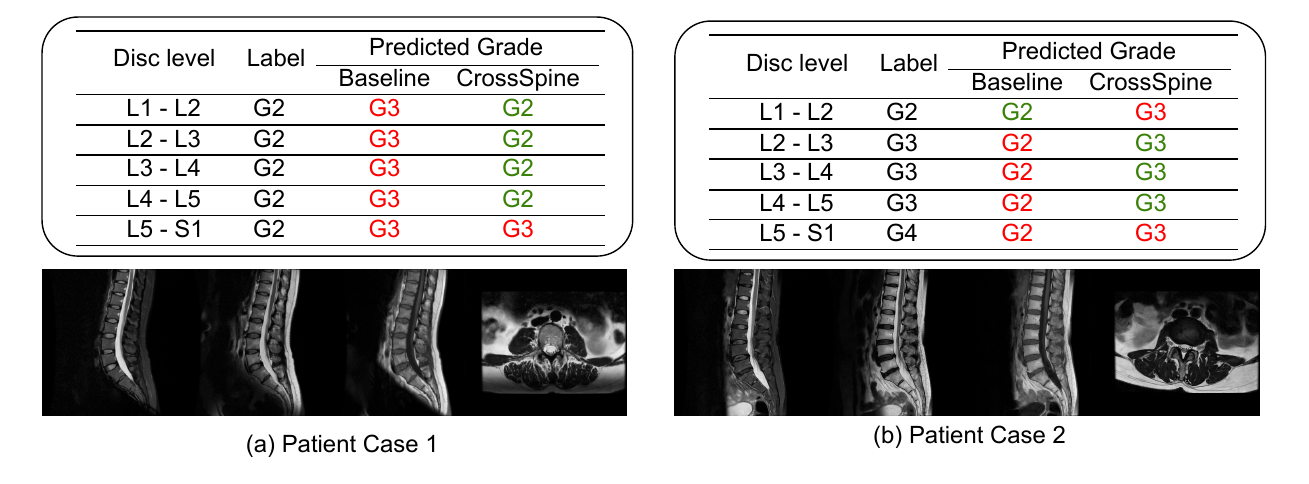}
    \caption{Visualization of prediction results on two patients. Green indicates correct predictions; red indicates errors.}
    \label{fig:result of two patient}
\end{figure}
To provide a qualitative assessment of our framework, Fig.~\ref{fig:result of two patient} visualizes the prediction results for two patients. As observed, the Baseline model frequently misclassifies the Pfirrmann grades, a limitation likely attributable to its reliance on single-stream features without anatomical context. In comparison, CrossSpine demonstrates superior performance by accurately aligning with the ground truth labels in most cases. This improvement confirms that integrating the cross-sequence attention mechanism and IVD-aware priors enables the model to effectively capture subtle degenerative patterns and anatomical dependencies that are overlooked by the baseline approach. In Case (b) shown in Fig.~\ref{fig:result of two patient}, the prediction pattern reveals distinct behaviors between the models. The Baseline model correctly identifies only the first disc, misclassifying all subsequent levels by consistently assigning the majority class (Grade 2). CrossSpine misclassifies only the L1--L2 and L5--S1 levels while successfully retrieving the correct grades for all intermediate discs. This highlights the effectiveness of the IVD-aware classification module in capturing level-specific degeneration context beyond simple majority voting.

Overall, these qualitative results demonstrate that integrating multi-scale feature fusion with anatomical context allows CrossSpine to outperform standard CNN backbones in Pfirrmann grading.
\begin{table}[H]
    \centering
    \caption{Performance comparison of DenseNet-121: Baseline (No IVD) vs. CrossSpine (With IVD).}
    \label{tab:experiment_results}
    \begin{tabular}{lccc}
        \toprule
        \textbf{Method} & \textbf{Macro F1} & \textbf{Mean AUROC} & \textbf{Mean AUPRC} \\
        \midrule
        Baseline (No IVD) & 0.1917 & 0.6187 & 0.2772 \\
        CrossSpine & \textbf{0.4318} & \textbf{0.8391} & \textbf{0.5516} \\
        \bottomrule
    \end{tabular}
\end{table}

Table~\ref{tab:experiment_results} illustrates the performance gap between the standard baseline model and our proposed CrossSpine framework. CrossSpine achieved a Macro F1 score of 0.4318 compared to the baseline of 0.1917, a relative improvement exceeding 125\%. The Mean AUROC improved from 0.6187 to 0.8391, while the Mean AUPRC nearly doubled from 0.2772 to 0.5516. This consistent gain across metrics indicates that integrating multiscale features and anatomical context enables the model to learn more robust structural representations than the baseline CNN.

\subsection{Backbone Architecture Analysis}
\begin{table}[H]
    \centering
    \caption{Performance comparison of representative backbones within the proposed framework. DenseNet-121 yields the highest consistency across F1, AUROC, and AUPRC.}
    \label{tab:backbone_comparison}
    \setlength{\tabcolsep}{6pt}
    \begin{tabular}{lccc}
        \toprule
        \textbf{Backbone} & \textbf{Macro F1} & \textbf{Mean AUROC} & \textbf{Mean AUPRC} \\
        \midrule
        ResNet-34       & 0.3992 & 0.6299 & 0.3634 \\
        EfficientNet-B0 & 0.4259 & 0.8364 & 0.5360 \\
        \textbf{DenseNet-121}    & \textbf{0.4318} & \textbf{0.8391} & \textbf{0.5516} \\
        \bottomrule
    \end{tabular}
\end{table}
To ensure a robust evaluation, we conducted preliminary experiments with representative architectures from three prominent CNN families: ResNet-34~\cite{he2016deep}, EfficientNet-B0~\cite{tan2019efficientnet}, and DenseNet-121~\cite{huang2017densely}. These variants were selected to balance computational efficiency with classification capability.

We evaluated these backbones within our proposed Multiscale IVD-aware framework. As detailed in Table~\ref{tab:backbone_comparison}, DenseNet-121 demonstrated superior capability across all prioritized metrics. It achieved the highest Macro F1 score (0.4318) and Mean AUROC (0.8391). While EfficientNet-B0 showed competitive performance, DenseNet-121 provided the best trade-off for capturing degenerative features. Consequently, DenseNet-121 was selected as the primary backbone for all subsequent ablation studies.

\subsection{Ablation Study}

To investigate the contributions of our proposed components, we performed an ablation study using the DenseNet-121 backbone, analyzing the impact of Multiscale and IVD-Aware modules as shown in Table~\ref{tab:ablation_study}. The results show that anatomical awareness is critical. Adding IVD-aware classification alone nearly doubled the baseline Macro~F1 (from 0.1917 to 0.3689). Similarly, Multiscale improved Macro~F1 to 0.3825. Integrating both modules yielded the best performance, achieving Mean~AUROC of 0.8391 and increasing Mean~AUPRC by about 99\% (from 0.2772 to 0.5516). This confirms that combining multi-view features with position awareness delivers the most accurate results.

\begin{table}[htbp]
    \centering
    \caption{Ablation study on the impact of the Multi-scale architecture and IVD-Aware classification mechanism. (Base: DenseNet-121).}
    \label{tab:ablation_study}
    \setlength{\tabcolsep}{8pt}
    \begin{tabular}{cc|ccc}
        \toprule
        \textbf{Multiscale} & \textbf{IVD-Aware} & \textbf{Macro F1} & \textbf{Mean AUROC} & \textbf{Mean AUPRC} \\
        \midrule
        - & - & 0.1917 & 0.6187 & 0.2772 \\   
        - & \checkmark & 0.3689 & 0.7865 & 0.4251 \\   
        \checkmark & - & 0.3825 & 0.8083 & 0.4785 \\   
        \checkmark & \checkmark & \textbf{0.4318} & \textbf{0.8391} & \textbf{0.5516} \\ 
        \bottomrule
    \end{tabular}
\end{table}

\section{Conclusion}
This study introduces CrossSpine, a framework that employs a cross-sequence attention mechanism to fuse multi-planar MRI representations. The release of a new dataset, combined with a model-agnostic IVD-aware classification strategy, demonstrates that incorporating anatomical priors consistently enhances diagnostic performance. These contributions establish a robust benchmark for multi-modal medical imaging and underscore the significance of anatomical context in automated spine analysis.

\section{Acknowledgement}
This research is supported by the Ben Dam Me Award Fund, the Vietnam Young Talent Support Fund, and the Number One Brand, Tan Hiep Phat Group.

\bibliographystyle{splncs04}

\bibliography{refs}       

@article{pfirrmann2001magnetic,
  title={Magnetic resonance classification of lumbar intervertebral disc degeneration},
  author={Pfirrmann, Christian WA and Metzdorf, Alexander and Zanetti, Marco and Hodler, Juerg and Boos, Norbert},
  journal={spine},
  volume={26},
  number={17},
  pages={1873--1878},
  year={2001},
  publisher={LWW}
}

@article{hartvigsen2018low,
  title={What low back pain is and why we need to pay attention},
  author={Hartvigsen, Jan and Hancock, Mark J and Kongsted, Alice and Louw, Quinette and Ferreira, Manuela L and Genevay, St{\'e}phane and Hoy, Damian and Karppinen, Jaro and Pransky, Glenn and Sieper, Joachim and others},
  journal={The Lancet},
  volume={391},
  number={10137},
  pages={2356--2367},
  year={2018},
  publisher={Elsevier}
}

@article{adams2006intervertebral,
  title={What is intervertebral disc degeneration, and what causes it?},
  author={Adams, Michael A and Roughley, Peter J},
  journal={Spine},
  volume={31},
  number={18},
  pages={2151--2161},
  year={2006},
  publisher={LWW}
}

@inproceedings{he2016deep,
  title={Deep residual learning for image recognition},
  author={He, Kaiming and Zhang, Xiangyu and Ren, Shaoqing and Sun, Jian},
  booktitle={Proceedings of the IEEE conference on computer vision and pattern recognition},
  pages={770--778},
  year={2016}
}

@inproceedings{huang2017densely,
  title={Densely connected convolutional networks},
  author={Huang, Gao and Liu, Zhuang and Van Der Maaten, Laurens and Weinberger, Kilian Q},
  booktitle={Proceedings of the IEEE conference on computer vision and pattern recognition},
  pages={4700--4708},
  year={2017}
}

@inproceedings{tan2019efficientnet,
  title={Efficientnet: Rethinking model scaling for convolutional neural networks},
  author={Tan, Mingxing and Le, Quoc},
  booktitle={International conference on machine learning},
  pages={6105--6114},
  year={2019},
  organization={PMLR}
}

@article{dosovitskiy2020image,
  title={An image is worth 16x16 words: Transformers for image recognition at scale},
  author={Dosovitskiy, Alexey},
  journal={arXiv preprint arXiv:2010.11929},
  year={2020}
}

@article{natalia2024lumbar,
  title={Lumbar spine MRI annotation with intervertebral disc height and Pfirrmann grade predictions},
  author={Natalia, Friska and Sudirman, Sud and Ruslim, Daniel and Al-Kafri, Ala},
  journal={PLoS One},
  volume={19},
  number={5},
  pages={e0302067},
  year={2024},
  publisher={Public Library of Science San Francisco, CA USA}
}

@article{vaswani2017attention,
  title={Attention is all you need},
  author={Vaswani, Ashish and Shazeer, Noam and Parmar, Niki and Uszkoreit, Jakob and Jones, Llion and Gomez, Aidan N and Kaiser, {\L}ukasz and Polosukhin, Illia},
  journal={Advances in neural information processing systems},
  volume={30},
  year={2017}
}

@article{russakovsky2015imagenet,
  title={Imagenet large scale visual recognition challenge},
  author={Russakovsky, Olga and Deng, Jia and Su, Hao and Krause, Jonathan and Satheesh, Sanjeev and Ma, Sean and Huang, Zhiheng and Karpathy, Andrej and Khosla, Aditya and Bernstein, Michael and others},
  journal={International journal of computer vision},
  volume={115},
  number={3},
  pages={211--252},
  year={2015},
  publisher={Springer}
}

@inproceedings{eigen2015predicting,
  title={Predicting depth, surface normals and semantic labels with a common multi-scale convolutional architecture},
  author={Eigen, David and Fergus, Rob},
  booktitle={Proceedings of the IEEE international conference on computer vision},
  pages={2650--2658},
  year={2015}
}

@book{goodfellow2016deep,
  title={Deep Learning},
  author={Goodfellow, Ian and Bengio, Yoshua and Courville, Aaron},
  publisher={MIT Press},
  year={2016},
  address={Cambridge, MA, USA}
}

@article{nguyen2025temporal,
  title={Temporal variational autoencoder model for in-hospital clinical emergency prediction},
  author={Nguyen, Trong-Nghia and Kim, Soo-Hyung and Kho, Bo-Gun and Do, Nhu-Tai and Iyortsuun, Ngumimi-Karen and Lee, Guee-Sang and Yang, Hyung-Jeong},
  journal={Biomedical Signal Processing and Control},
  volume={100},
  pages={106975},
  year={2025},
  publisher={Elsevier}
}

@article{asif2025advancements,
  title={Advancements and prospects of machine learning in medical diagnostics: unveiling the future of diagnostic precision},
  author={Asif, Sohaib and Wenhui, Yi and ur-Rehman, Saif- and ul-ain, Qurrat- and Amjad, Kamran and Yueyang, Yi and Jinhai, Si and Awais, Muhammad},
  journal={Archives of Computational Methods in Engineering},
  volume={32},
  number={2},
  pages={853--883},
  year={2025},
  publisher={Springer}
}

@inproceedings{kowlagi2023stronger,
  title={A stronger baseline for automatic Pfirrmann grading of lumbar spine MRI using deep learning},
  author={Kowlagi, Narasimharao and Nguyen, Huy Hoang and McSweeney, Terence and Saarakkala, Simo and M{\"a}{\"a}tt{\"a}, Juhani and Karppinen, Jaro and Tiulpin, Aleksei},
  booktitle={2023 IEEE 20th International Symposium on Biomedical Imaging (ISBI)},
  pages={1--5},
  year={2023},
  organization={IEEE}
}

@article{lundervold2019overview,
  title={An 
  of deep learning in medical imaging focusing on MRI},
  author={Lundervold, Alexander Selvikv{\aa}g and Lundervold, Arvid},
  journal={Zeitschrift f{\"u}r Medizinische Physik},
  volume={29},
  number={2},
  pages={102--127},
  year={2019},
  publisher={Elsevier},
}

\end{document}